# Method of Tibetan Person Knowledge Extraction

Sun Yuan[*,1,2], Zhu Zhen[1,2]

[1]*School of Information Engineering, Minzu University of China, Beijing, 100081, P.R. China*

[2] *Minority Languages Branch, National Language Resource and Monitoring Research Center*

**Abstract:** Person knowledge extraction is the foundation of the Tibetan knowledge graph construction, which provides support for Tibetan question answering system, information retrieval, information extraction and other researches, and promotes national unity and social stability. This paper proposes a SVM and template based approach to Tibetan person knowledge extraction. Through constructing the training corpus, we build the templates based the shallow parsing analysis of Tibetan syntactic, semantic features and verbs. Using the training corpus, we design a hierarchical SVM classifier to realize the entity knowledge extraction. Finally, experimental results prove the method has greater improvement in Tibetan person knowledge extraction.

## 1. INTRODUCTION

Explosive growth of Web content is making the study of social network on Web from the structure analysis to the content analysis, Knowledge Graph is becoming a hot research of Natural Language Processing in the age of big data [1]. According to the survey results from the National Language Resource Monitoring and Research Center of Minzu University of China in 2013: the number of China ethnic minority language websites is 1,031, which include 655 Uyghur websites, 104 Tibetan websites and 102 Mongolian websites.

Knowledge graph with full and complete knowledge system for information retrieval, question answering system, construction of knowledge base and other areas of study provides resources and support. Nodes of Knowledge Graph represent entities and concepts, and connecting edges represent various semantic relations between entities and concepts, and entity knowledge extraction is one of the main research contents.

At present, the existing knowledge graph only provides the relevant knowledge of English, Chinese or France, such as Google (more than 5.7 billion entities, 18 billion connections), DBpedia (more than 1,900 million entities. 1 billion relations), Wiki-links (40 million to disambiguate the relations), Wolframalpha (10 trillion), Probase (more than 265 million entities), Baidu, Sogou, etc [2]. Minority language knowledge graph research is just beginning.

For example, when we search "ཏཱ་ལའི་བླ་མ (the Dalai Lama)", Google has 64,100 results. And when we search "རྒྱལ་བ་རིན་པོ་ཆེ (Jiawa Rinpoche)", Google has 586,000 results. In Tibetan, commonly known ཏཱ་ལའི་བླ་མ (the Dalai Lama) as རྒྱལ་བ་རིན་པོ་ཆེ (Jiawa Rinpoche), and the current search engines do not show the relation between them.

In addition, all the search results are based on keywords, without the structure of knowledge representation. If we have the semantic link between entity and entity, then we will get more comprehensive information to realize the deep information mining.

## 2. RELATED WORKS

The key problem of entity knowledge extraction is the entity relation extraction. Tibetan entity knowledge extraction techniques include the training corpus construction, text representation, entity relation extraction. The present situation is as follows.

In training corpus construction, entity knowledge extraction method based on machine learning requires a certain scale of training corpus, and the corpus using artificial mark spends a lot of time and manpower. In English and Chinese, training corpus mainly comes from ACE, SemEval and natural annotation corpus [3-10]. However, most of this corpus aims at English and Chinese. For Tibetan entity relation extraction, there is no ready-made training corpus can be directly obtained. In recent years, some researchers try to use the rich resource language to help the less resource language, and build the training corpus. Kim [11] put forward a cross-language relation extraction method based on label mapping, which applied the training model to the source language in parallel corpus, and mapped the identified high reliability examples to the target language, finally obtained the training corpus to the entity relation of target language. Kim and Lee [12] further used a semi-supervised learning algorithm, which mapped more contextual information in the source language into the target language through iterative method, and improved the quantity and quality of training corpus. Yanan Hu [13] used machine translation to realize the relation example transformation from the source language to target language, and helped the less resource language to extract semantic relations.

*No.27, Zhongguancun South Street, Haidian District, Beijing, 100081, P.R. China; Tel: 8610-68930880. E-mail: tracy.yuan.sun@gmail.com

In the aspect of text representation, the most common method of word representation is the One-hot Representation. In this method, word is expressed as a very long vector and the dimension is vocabulary size. The problem of One-hot Representation is dimension disaster. Meanwhile, there is the phenomenon of isolation between words. Aiming at this problem, some researchers combined semantics knowledge base, such as WordNet, HowNet, or Distributed Representation to improve the accuracy of entity relation description [14-22]. However, Tibetan semantic knowledge base construction is not well developed. At present, there is not a practical Tibetan semantic knowledge base system.

In terms of entity relation extraction, the typical methods include feature vector [23,24] and kernel function [25,26]. Feature vector method includes the maximum entropy model [24] and support vector machine (SVM) [28,29]. The main focus of this method is how to obtain all kinds of effective features. Qin Deng [30] introduced the lexical semantic matching technology into Chinese entity relation extraction based on pattern matching technology, and compared the performance between the general pattern matching technology and lexical semantic pattern matching technology. Jifa Jiang [31] proposed a bootstrap acquisition method based on binary relation mode. Weiru Zhang [27] proposed a method based on Wikipedia and mode clustering, which has highly accuracy in Chinese entity relation extraction.

Different from the method based on feature vector, the method based on kernel function does not need to construct high dimensional feature vector space. Through calculating the similarity between two discrete objects (such as syntax tree structure), this method uses the parsing structure tree to realize the classification. Zelenko [32] introduced the kernel function method into the relation extraction, which defined the kernel function on the basis of shallow parsing and designed a dynamic programming algorithm to extract entity semantic relations, finally got good results in 200 news texts. Culotta [33] transformed the parse tree into the dependency tree through some transformation rules, and increased features of the part of speech, entity types, phrases, WordNet, used the SVM classifier to relation extraction. Through testing ACE RDC 2003 data, the F value achieved 45.8%. Bunescu [34] further proposed the kernel function based on the shortest path dependence tree, and F value achieved 52.5% on the ACE RDC 2003 data, but the recall rate is lower. Zhang [35] designed a composite convolution kernel function tree for relation extraction. The method combined convolution kernel function with linear kernel function, fully considering the influence of the semantic relations. The F value reached 70.9% and 72.1% respectively on the ACE 2003 and ACE 2004 data. Kebin Liu [36] has realized the automatic extraction system based on the improved semantic sequence kernel function and combined with KNN machine learning algorithm to entity relation.

At present, Tibetan display output technology, coding technology, input technology, text processing technology, the homepage manufacture technology obtained the very good solution, however in the study of sentence and chapter level, Tibetan is still in the initial stage. Therefore, some entity relation extraction method in English and Chinese cannot be directly applied to Tibetan.

In this condition, this paper proposes a SVM and template based approach to Tibetan person knowledge extraction. The main work of this paper is in the following.

(1) Constructing the training corpus.

(2) Building the templates based the shallow parsing analysis of Tibetan syntactic, semantic features and verbs.

(3) Using the training corpus, we design a hierarchical SVM classifier to realize the entity knowledge extraction.

The research of Tibetan person knowledge extraction is the foundation of Tibetan knowledge graph construction. It provides support for Tibetan question answering system, information retrieval, information extraction and other researches, and promotes national unity and social stability.

## 3. TRAINING CORPUS AND TEMPLATE CONSTRUCTION

Firstly, we crawl data from some websites, such as Wikipedia, Comba Media Net. Through word segmentation, POS tagging and entity recognition, we use templates and SVM to realize the entity knowledge extraction.

### 3.1 Training corpus

We use the "entity, attribute and value" of the existing information box in Chinese from Wikipedia to obtain the Chinese sentences with the entity and the attribute. Based on the correspondence between entities in the parallel sentence of Tibetan and Chinese, we construct some training corpus of Tibetan entity knowledge extraction.

Meanwhile, based on the shallow parsing analysis of Tibetan syntactic, semantic features and verbs, we get the templates and the main features of Tibetan entity knowledge extraction, shown in Fig. (1).

### 3.2 Template building

Tibetan is a predicate behind language and the verb is the core of the sentence. Meanwhile, Tibetan auxiliary can clearly indicate the semantic structure of sentence, shown in Table 1.

Therefore, the feature selection mainly includes Tibetan case makers, verbs and other substantives.

Example 1: སྒྲོལ་མ་1988ལོར་སྐྱེས།(Zhuo Ma was born in 1988.)

Template 1:<name/nh><time/t>(སུ|རུ|ར|ལ|ཏུ|དུ|ནར)/u སྐྱེས།/v[born]

Example 2: སྒྲོལ་དཀར་གྱི་ཕ་ཡུལ་ནི་མཚོ་སྔོན་དུ་ཡིན་(Zhuo Ga's hometown is Qinghai. )

Template 2: <name/nh>(གི|གྱི|ཀྱི|འི|ཡི)/u ཕ་ཡུལ/n[hometown] ནི་/r <place name/np>(སུ|རུ|ར|ལ|ཏུ|དུ|ན)/u ཡིན།/v

Example 3： ཚེ་དབང་གི་སྐྱེས་སྐར་ནི་1988ལོའི་ཟླ10བའི་ཚེས1ཉིན་ཡིན།(Tsewang's birthday is October 1, 1988.)

Template 3:<name/nh>(གི|གྱི|ཀྱི|འི|ཡི)/u སྐྱེས་སྐར་/n[birthday] ནི་/r<time/t> ཡིན།/v

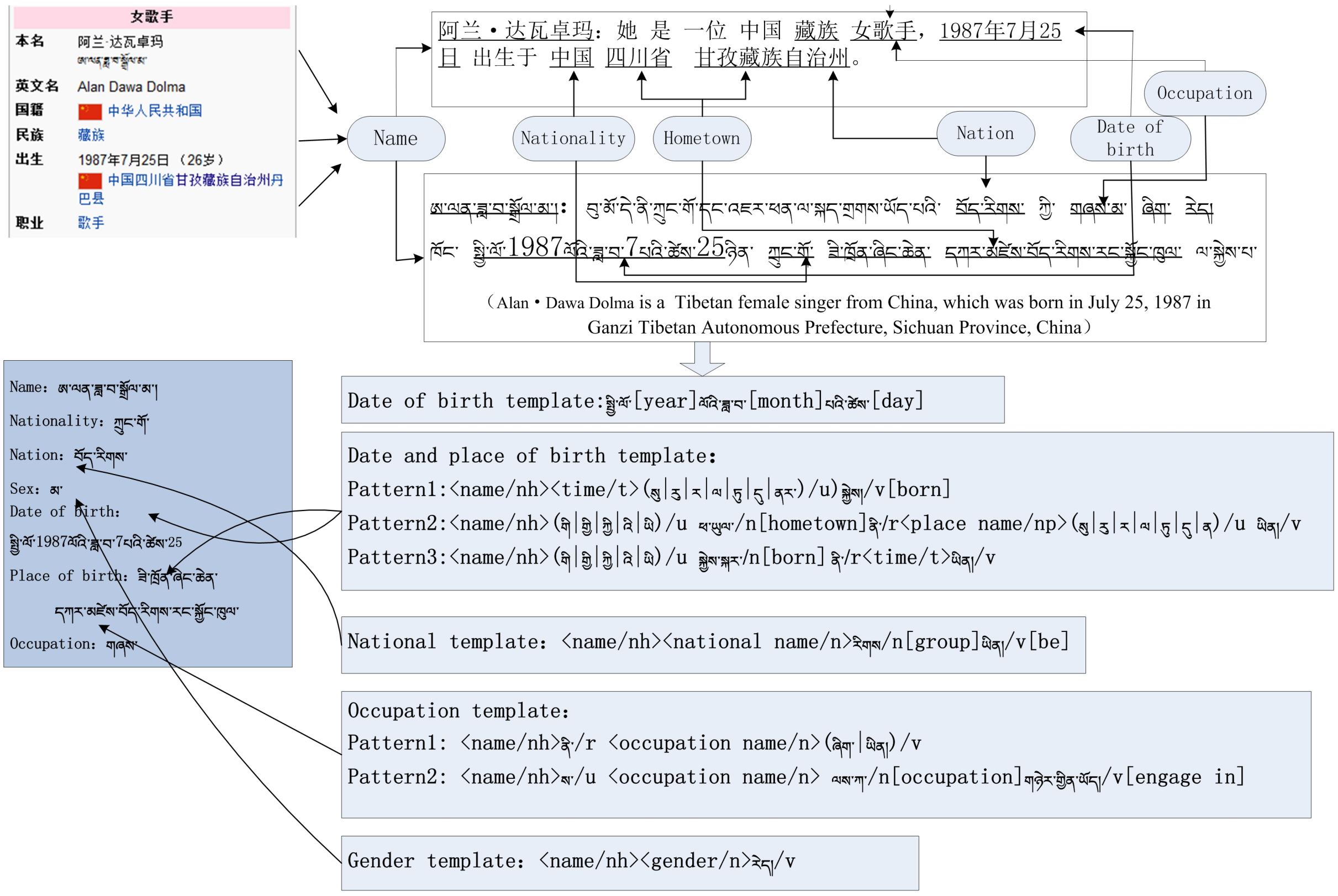

**Fig. (1).** Tibetan Entity Relation Template Construction.

## 4. HIERARCHICAL CLASSIFICATION BASED ON SVM

Although the method based on feature template can obtain higher accuracy in specific test corpus, it requires a lot of manual works and cannot extract the corpus which is not covered. Therefore, we adopted the SVM method based on feature vector, and designed a hierarchical classifier.

### 4.1 Feature selection

In this paper, the selected features include keywords, tagging combination feature and words around entities.

(1) Keywords

Keywords are a noun or verb with high frequency and strong distinction. Most of these features are extracted from the templates. Although keywords are not many, these words have a strong distinction in a certain attribute class.

(2) Tagging combination feature

The part-of-speech tagging is an important feature. But not every tag can be used as a feature vector, because many tags are not discrimination. Therefore, we use tagging combination features, which get better classification effect. For example, the combination of time tag "/t" + auxiliary "/k" + "/ v " has great help for identifying birth attribute.

(3) Words around entities

The feature of word around entity includes parts of speech tagging and named entity marking. We consider that markers which are nearer to the entity are more important. Therefore, we select two words forward entity and one word backward entity.

### 4.2 Hierarchical classifier construction

SVM was originally designed to solve the problem of binary classification, but entity attribute extraction is the multi classification problem. For example, person attributes can be divided into birth, birth place, gender and other categories. So the key problem is to design the high performance classifier.

At present, there are two types of classifier:

(1) One-to-one classifier. If there are $k$ categories, we need to build $k(k-1)/2$ classifiers, and then calculate $k(k-1)/2$ times and get the cumulative weight .The largest cumulative value is the category.

(2) One-to-many classifier. This method need to build $k$ classifiers if there are $k$ categories, then we need to calculate $k/2$ times to forecast each attribute.

Comparing the two methods, the first method is better than the second, but the number of classifiers is too much. When the categories are so much, the applicability is poor.

Therefore, this paper designs a hierarchical classifier. The method combines the advantages of the two traditional methods, shown in Fig. (2).

**Table 1. Grammatical and Semantics Functions of Tibetan Case Markers**

| Type | Example of Case Marker | Contained Subtype | Function of Syntax and Semantics |
|---|---|---|---|
| Nominative case | གིས、 ཀྱིས、 གྱིས、 ཡིས、 འིས | Apply case | Indicate the agent of action |
| | | Tools case | Indicate the tools and the way for action |
| Possessive phrases case | གི、 གྱི、 ཀྱི、 འི、 ཡི | | Indicate possessive phrases |
| Pull case | སུ、 ཏུ、 ར、 ལ、 དུ、 རུ、 ན | Occupation case | Indicate the object, location and so on |
| | | For case | Indicate the person or action benefited |
| | | Depend case | Indicate the relation of interdependence or location |
| | | Community case | Indicate things situation |
| | | Time case | Indicate the time happened |
| From case | ནས、 ལས | | Indicate the source of action or situation |

In the same level, we use one-to-one method. Meanwhile, we use language rules to build a fast track, which has certain advantages in the classification results, speed and the number of classifiers.

(1) The filter: Before entering the hierarchical classifier, it is necessary to make a selection in the corpus. The filter will eliminate the sentence which has no entities. To a certain extent, it can reduce the workload of the hierarchical classifier and improve the efficiency.

(2) Layer by layer to down: After entering the hierarchical classifier system, the standard classification model is to start from the first layer to the bottom layer. The intermediate classifier will eliminate some of the unrelated data. This step is very important to deal with a large number of negative samples in attribute extraction.

(3) Multi classification in the same layer: After the hierarchical classification, there are not many categories to deal with using each multi classifier. The number of classifiers is $N_{sum} = \sum_{i=1}^{n} p_i(p_i - 1)/2$ , $p_i$ is the number of categories in each classifier. It solves the problem of classifications are too small and keep the high accuracy of one-to-one method.

(4) Fast track: we design a fast track based on entity-attribute tagging characteristics, which can effectively improve the classification effect and speed, because the entity attribute often has obvious distinction in the attribute extraction. For example, when the second entity is time, it only appears the birth date attribute but not a father or a birth place. So it can directly jump to the category of birth date through the fast track.

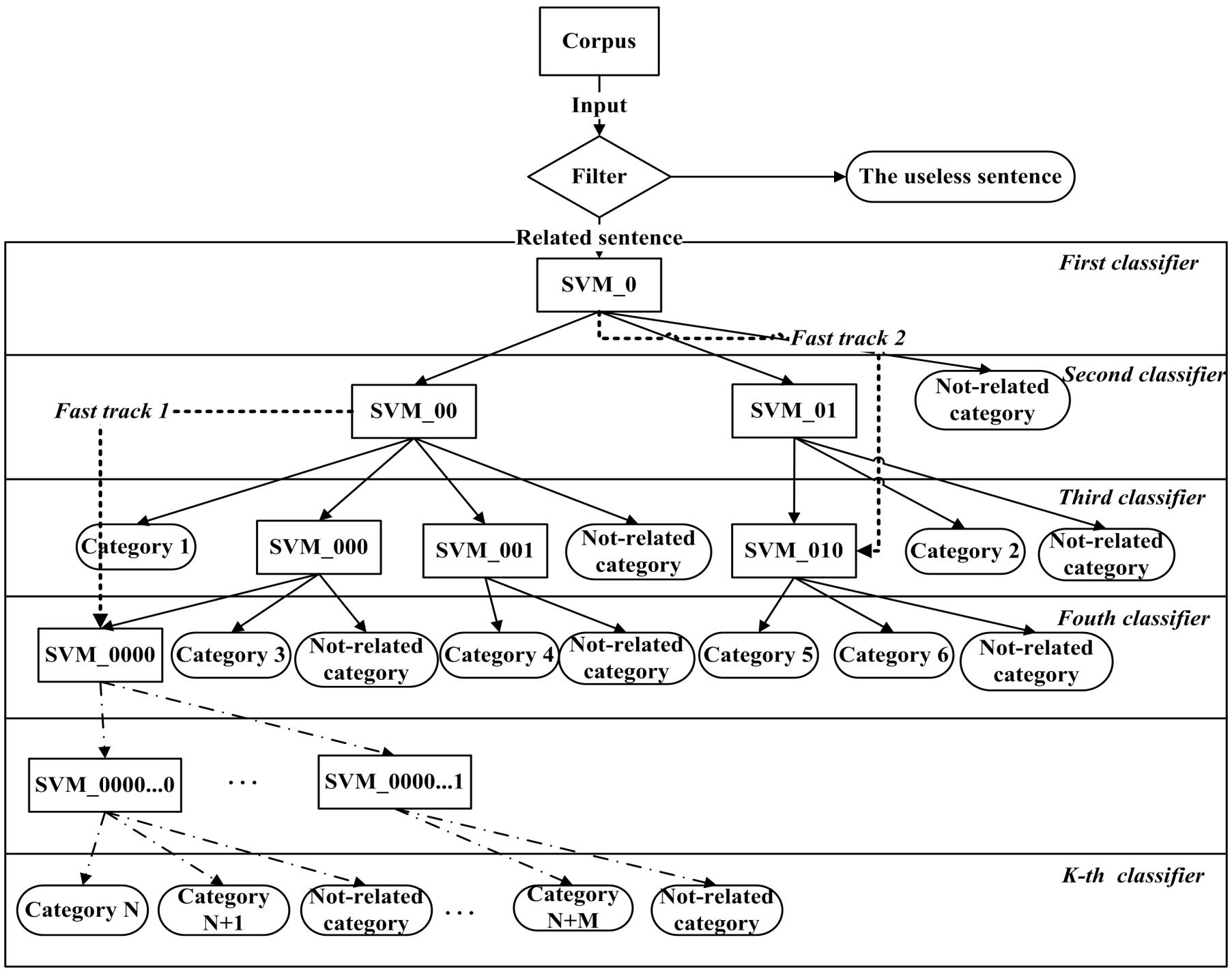

**Fig. (2).** Hierarchical Classifier.

## 5. THE EXPERIMENTAL RESULTS AND ANALYSIS

### 5.1 Corpus sources

In this paper, corpus data is from seven Tibetan website, as shown in Table 2. We mainly process the following four entity relations: Name-Birth date, Name- Birth place, Name-Father, Name-Mother.

**Table 2. Corpus**

| Corpus source | URL |
|---|---|
| Wikipedia (Tibetan) | http://bo.wikipedia.org |
| China Tibetan Middle School Net | http://www.tibetanms.cn/ |
| Comba Media Net | http://ti.kbcmw.com |
| Himalaya BON (Tibetan) | http://old.himalayabon.com/ |
| AMDO Tibetan | http://www.amdotibet.cn/ |
| HIMALAYABON | http://www.himalayabon.com/ |
| Jetsongkhapa Net | http://bo.jetsongkhapa.org/ |

We select 2,400 sentences that contain entity-attributes from a large number of web pages. Among them, the 1,975 sentences are contained above four entity-attributes and 425 sentences are for other entity-attributes. We select 1,600 sentences as training corpus, and 800 sentences as test corpus.

### 5.2 The experimental analysis and evaluation

Firstly, we test the template method on the open set (800 sentences), which contain 846 attributes. The experimental results are shown in Table 3.The experimental results show that the performance of template method used in open set is not high. The main reason is that this method lacks learning ability and must be constructed through artificial participation. Although performance will gradually increase by generalization and correction templates, too much human intervention becomes the bottleneck of the method.

Secondly, we use the SVM hierarchical classifier to extract entity relations, the experimental results are shown in Table 4. Compared with the template method, SVM method improves the recall rate of entity-attribute extraction but accuracy is not improved obviously. The main reason is that the results using SVM can get good performance on the non-obvious classification by diversifying the feature vectors, but there are some mistakes in some clear classification, because of training corpus insufficient.

Finally, we use the method based on template and SVM method. The experimental results are shown in Table 5.

Table 5 shows the performance of using template and SVM is obviously higher than only using any one method.

**Table 3. Results on the Open Set Using Template Method**

| Category | Number | | | Percentage | | |
|---|---|---|---|---|---|---|
| | *Total* | *Identified* | *Correct* | *P* | *R* | *F1* |
| Birth Day | 219 | 162 | 91 | 56.17% | 41.55% | 47.77% |
| Birth Place | 223 | 168 | 78 | 46.43% | 34.98% | 39.90% |
| Father | 184 | 144 | 73 | 50.69% | 39.67% | 44.51% |
| Mother | 220 | 171 | 87 | 50.88% | 39.55% | 44.50% |

**Table 4. Results on the Open Set Using SVM Method**

| Category | Number | | | Percentage | | |
|---|---|---|---|---|---|---|
| | *Total* | *Identified* | *Correct* | *P* | *R* | *F1* |
| Birth Day | 219 | 202 | 103 | 50.99% | 47.03% | 48.93% |
| Birth Place | 223 | 211 | 94 | 44.55% | 42.15% | 43.32% |
| Father | 184 | 176 | 83 | 47.16% | 45.11% | 46.11% |
| Mother | 220 | 208 | 101 | 48.56% | 45.91% | 47.20% |

**Table 5. Results on the Open Set Using Template and SVM Method**

| Category | Number | | | Percentage | | |
|---|---|---|---|---|---|---|
| | *Total* | *Identified* | *Correct* | *P* | *R* | *F1* |
| Birth Day | 219 | 201 | 131 | 65.17% | 59.82% | 62.38% |
| Birth Place | 223 | 209 | 133 | 63.64% | 59.64% | 61.57% |
| Father | 184 | 161 | 108 | 67.08% | 58.70% | 62.61% |
| Mother | 220 | 201 | 128 | 63.68% | 58.18% | 60.81% |

## 6. CONCLUSION

In this paper, we propose a SVM and template based approach to Tibetan person knowledge extraction. And the experimental results show the method achieves good performance. In the following work, we will increase the training corpus, and improve this method. Then we will use CRF, word embedding method to the Tibetan entity knowledge extraction, and give the comparison results.

## CONFLICT OF INTEREST

The author confirms that this article content has no conflict of interest.

## ACKNOWLEDGMENTS

This work is supported by National Nature Science Foundation (No. 61501529, No. 61331013), National Language Committee Project (No. YB125-139, ZDI125-36), and Minzu University of China Scientific Research Project (No. 2015MDQN11, No. 2015MDTD14C).